\documentclass[twoside,11pt]{article}

\usepackage{blindtext}

\usepackage[abbrvbib, preprint]{jmlr2e}

\usepackage{amssymb}
\usepackage{mathtools}
\usepackage{fancybox}
\usepackage{multirow}
\usepackage{paralist}
\usepackage{subfigure}
\RequirePackage{algorithm}
\RequirePackage{algorithmic}
\newtheorem{lem}{Lemma}
\newtheorem{ass}{Assumption}
\newtheorem{defi}{Definition}
\newcommand{\Norm}[1]{\left\|#1\right\|}
\usepackage{makecell} 

\def \E {\mathbb{E}}
\def \R {\mathbb{R}}

\def \v {\mathbf{v}}

\def \x {\mathbf{x}}
\def \y {\mathbf{y}}

\def \sign  {\operatorname{Sign}}

\usepackage{pifont}

\usepackage{xcolor}  
\usepackage{bm}

\def \y {\mathbf{y}}
\def \E {\mathbb{E}}
\def \x {\mathbf{x}}
\def \v {\mathbf{v}}

\def \R {\mathbb{R}}

\def \m {\mathbf{m}}

\ShortHeadings{}{Jiang and Zhang}
\firstpageno{1}

\begin{document}

\title{Convergence Analysis of the Lion Optimizer \\
in Centralized and Distributed Settings}

\author{\name Wei Jiang \email jiangw@lamda.nju.edu.cn \\
       \addr National Key Laboratory for Novel Software Technology, Nanjing University, China
       \AND
       \name Lijun Zhang \email zhanglj@lamda.nju.edu.cn \\
       \addr National Key Laboratory for Novel Software Technology, Nanjing University, China \\ School of Artificial Intelligence, Nanjing University, China}

\editor{My editor}

\maketitle

\begin{abstract}
In this paper, we analyze the convergence properties of the Lion optimizer. First, we establish that the Lion optimizer attains a convergence rate of $\mathcal{O}(d^{1/2}T^{-1/4})$ under standard assumptions, where $d$ denotes the problem dimension and $T$ is the iteration number. To further improve this rate, we introduce the Lion optimizer with variance reduction, resulting in an enhanced convergence rate of $\mathcal{O}(d^{1/2}T^{-1/3})$. We then analyze in distributed settings, where the standard and variance reduced  version of the distributed Lion can obtain the convergence rates of $\mathcal{O}(d^{1/2}(nT)^{-1/4})$ and $\mathcal{O}(d^{1/2}(nT)^{-1/3})$, with $n$ denoting the number of nodes. Furthermore, we investigate a communication-efficient variant of the distributed Lion that ensures sign compression in both communication directions. By employing the unbiased sign operations, the proposed Lion variant and its variance reduction counterpart, achieve convergence rates of $\mathcal{O}\left( \max \left\{\frac{d^{1/4}}{T^{1/4}}, \frac{d^{1/10}}{n^{1/5}T^{1/5}} \right\} \right)$ and $\mathcal{O}\left( \frac{d^{1/4}}{T^{1/4}} \right)$, respectively.
\end{abstract}

\section{Introduction}
The Lion (evolved sign momentum) optimizer~\citep{chen2023symbolic}, introduced by Google through program search, is an efficient optimization algorithm known for its memory efficiency and strong generalization performance. It has demonstrated effectiveness across a broad range of tasks, including large language model fine-tuning, diffusion models, and vision-language contrastive training, among others. In this paper, we investigate the convergence properties of the Lion optimizer within the framework of stochastic optimization, formulated as:
\begin{align}\label{problem1}
\min_{\x \in \R^d} f(\x),
\end{align}
where $f: \R^d \to \R$ is a smooth and non-convex function. We assume that only noisy gradient estimates are available, denoted by $\nabla f(\x; \xi)$, where $\xi$ represents a random sample drawn from a stochastic oracle, such that $\E\left[\nabla f(\x; \xi)\right] = \nabla f(\x)$.

Although the Lion optimizer exhibits superior empirical performance, its theoretical properties have yet to be fully explored. Previous works~\citep{dong2024} have derived a convergence rate of $\mathcal{O}\left(d^{1/2}T^{-1/4} \right)$, but this result requires the additional assumption of function coerciveness. A similar result is also obtained by \cite{sfyraki2025}, though their analysis depends on large batch sizes and is derived in terms of the Frank-Wolfe gap for constrained problems. In contrast, we focus on the convergence to the stationary point for the unconstrained problem~(\ref{problem1}), where we establish the $\mathcal{O}\left(d^{1/2}T^{-1/4} \right)$ rate under standard assumptions. Furthermore, by incorporating variance reduction estimators, we show that the convergence rate of the Lion optimizer can be improved to $\mathcal{O}\left(d^{1/2}T^{-1/3} \right)$ with a minor modification. 

The study of the Lion optimizer is not limited to the centralized settings. However, existing literature on distributed settings typically requires strong additional assumptions to derive convergence guarantees. Notably, \cite{liu2024communication} analyze convergence properties for constrained problems under the assumption of bias correction, where it is assumed that the expectation of the stochastic gradient and its sign share the same sign. Additionally, \cite{10447045} derive convergence guarantees for the FedLion algorithm but impose the strong assumption of bounded heterogeneity. In contrast, we aim to provide theoretical guarantees for the distributed Lion optimizer under standard assumptions. Specifically, we establish a convergence rate of $\mathcal{O}\left(d^{1/2}(nT)^{-1/4} \right)$ for distributed Lion, where $n$ denotes the number of nodes in the distributed system. By further employing variance reduction estimators, we improve the rate to $\mathcal{O}\left(d^{1/2}(nT)^{-1/3} \right)$, matching the results obtained in the centralized settings.

Furthermore, we explore the communication-efficient version of the  distributed Lion optimizer. Although the original Lion optimizer uses the sign operation for updates, its distributed version can only ensure 1-bit sign compression in a single communication direction. By utilizing the sign operator in both the server and local nodes, we introduce a variant that guarantees 1-bit compression in both communication directions. For this method, we can derive convergence rates of $\mathcal{O}\left( \frac{d^{1/2}}{T^{1/2}} + \frac{d}{n^{1/2}} \right)$ and $\mathcal{O}\left( \frac{n^{1/2}}{T} + \frac{d}{n^{1/2}} \right)$ respectively. Additionally, by using the unbiased sign operations in both directions, we further improve the convergence rates to $\mathcal{O}\left( \max \left\{\frac{d^{1/4}}{T^{1/4}}, \frac{d^{1/10}}{n^{1/5}T^{1/5}} \right\} \right)$ and $\mathcal{O}\left( \frac{d^{1/4}}{T^{1/4}} \right)$ for the distributed Lion variant and its variance reduction counterpart.

In summary, the contributions of this paper are as follows:
\begin{itemize}
    \item We establish a convergence rate of $\mathcal{O}(d^{1/2}T^{-1/4})$ for the Lion optimizer under standard assumptions, without relying on the coerciveness assumption that has been used in the previous work~\citep{dong2024}. Furthermore, we improve this rate to $\mathcal{O}(d^{1/2}T^{-1/3})$ by employing the variance reduction technique.
    \item In distributed settings, we show that the distributed Lion optimizer and its variance reduced variant achieve convergence rates of $\mathcal{O}(d^{1/2}(nT)^{-1/4})$ and $\mathcal{O}(d^{1/2}(nT)^{-1/3})$, respectively, matching the rates obtained in centralized settings.
    \item We introduce highly communication-efficient variants of the distributed Lion, which ensure 1-bit sign compression in both communication directions. These variants can achieve convergence rates of $\mathcal{O}\left( \frac{d^{1/2}}{T^{1/2}} + \frac{d}{n^{1/2}} \right)$ and $\mathcal{O}\left( \frac{n^{1/2}}{T} + \frac{d}{n^{1/2}} \right)$. By employing the unbiased sign operations in both directions, we further improve the convergence rates to $\mathcal{O}\left( \max \left\{\frac{d^{1/4}}{T^{1/4}}, \frac{d^{1/10}}{n^{1/5}T^{1/5}} \right\} \right)$ and $\mathcal{O}\left( \frac{d^{1/4}}{T^{1/4}} \right)$, respectively.
\end{itemize}

\section{Related work}
This section reviews related work on theoretical analysis of the Lion optimizer, stochastic optimization and variance reduction, and sign-based methods.
\subsection{Theoretical analysis of the Lion optimizer}
The Lion optimizer~\citep{chen2023symbolic}, originally discovered by Google through symbolic program search, has shown superior performance in training large AI models, including large language models. It often outperforms AdamW in terms of both training speed and memory efficiency. Despite its empirical success, the theoretical foundations of Lion were not fully understood until recent studies. 
A notable contribution~\citep{chen2024lion} demonstrates that Lion operates as a constrained optimization algorithm, effectively minimizing a general loss function $f(x)$ subject to the constraint $\|x\|_\infty \leq 1/\lambda$. This is achieved through the incorporation of decoupled weight decay, where $\lambda$ represents the weight decay coefficient.

Further research~\citep{dong2024} establishes that Lion converges to Karush-Kuhn-Tucker (KKT) points at a rate of $\mathcal{O}(d^{1/2} T^{-1/4})$ for constrained optimization problems, where $d$ represents the problem dimension. This convergence result extends to the unconstrained case with the same rate, under the additional assumption of coerciveness of the objective function.
Subsequent work by \cite{sfyraki2025} highlights that Lion can be viewed as a special case of the Stochastic Frank-Wolfe algorithm. They establish a similar convergence guarantee in terms of the Frank-Wolfe gap by using large batch sizes.

In the context of distributed optimization, \cite{lioncub} integrate quantization techniques into the distributed Lion, although they do not provide theoretical guarantees for their approach. While some literature analyzes the convergence properties of distributed variants of Lion, these studies typically rely on additional assumptions.
For instance, \cite{liu2024communication} derive convergence guarantees for constrained problems with the assumption of bias correction, requiring that the expected values of the stochastic gradient and its sign align. Additionally, \cite{10447045} analyze the FedLion algorithm under the strong assumption of bounded heterogeneity.

\subsection{Stochastic optimization and variance reduction}
Stochastic optimization has been widely investigated in the machine learning community, focusing on studying convergence properties of stochastic algorithms. It is well known that the traditional stochastic gradient descent~(SGD) method can obtain a convergence rate of $\mathcal{O}\left(T^{-1/4} \right)$~\citep{SGD}, matching the lower bound under the standard smoothness assumption~\citep{Arjevani2019LowerBF}. By incorporating the past gradients into the current update, the momentum technique is introduced to accelerate SGD, which helps escape saddle points faster~\citep{Wang2020Escaping} and maintains the same convergence rate~\citep{SGDMc}.

To further improve the convergence rate, variance reduction techniques have been developed, which accelerate convergence by reducing variance in gradient estimations. Classical algorithms include SAG~\citep{DBLP:conf/nips/RouxSB12}, SVRG~\citep{NIPS:2013:Zhang,NIPS2013_ac1dd209}, and SARAH~\citep{arxiv.1703.00102}, which obtain improved convergence rate for convex or strongly convex functions. For the general non-convex function, the SPIDER algorithm improves the convergence of SGD to $\mathcal{O}\left( T^{-1/3} \right)$ under the average smoothness assumption. This rate can be further improved to $\mathcal{O}\left(n^{1/4} T^{-1/2} \right)$ for the finite-sum problem, where $n$ is the number of functions in the finite-sum structure. To avoid the huge batch size required in the SPIDER algorithm, \cite{cutkosky2019momentum} propose the STORM method, attaining the same $\mathcal{O}\left( T^{-1/3} \right)$ convergence rate with constant batch sizes.

\subsection{Sign-based optimization methods}
Sign-based optimization methods, such as SignSGD, have gained popularity due to their simplicity and communication efficiency, especially in distributed training scenarios. These methods update model parameters using only the sign of the gradient, reducing the amount of data transmitted between nodes. The convergence of signSGD methods has first been studied by \cite{pmlr-v80-bernstein18a, bernstein2018signsgd}, which obtain the convergence rate of $\mathcal{O}(d^{1/2}T^{-1/4})$, where $d$ is the dimension. To avoid the huge batch size or the unimodal symmetric noise assumption used in the previous analysis~\citep{pmlr-v80-bernstein18a, bernstein2018signsgd}, later literature~\citep{pmlr-v202-sun23l,jiang2025sign} uses the momentum technique and attains a similar convergence rate of signSGD under more standard assumptions. This rate is further improved to  $\mathcal{O}(d^{1/2}T^{-1/3})$ by \cite{NeurIPS:2024:Jiang:B}, using variance reduction estimators.

In distributed settings, sign-based methods with majority voting have also been widely investigated due to their communication efficiency. In the homogeneous environment, \cite{pmlr-v80-bernstein18a, bernstein2018signsgd} show that signSGD with majority voting obtains the $\mathcal{O}(d^{1/2}T^{-1/4})$ convergence rate. For more challenging heterogeneous environments, MV-sto-signSGD-SIM~\citep{pmlr-v202-sun23l} attain the convergence rate of $\mathcal{O}\left(dT^{-1/4} + {d}n^{-1/2}\right)$, where $n$ is the number of nodes. This rate is further improve to $\mathcal{O}\left(d^{1/2}T^{-1/2} + dn^{-1/2}\right)$ by \cite{jiang2025sign}. To ensure that that gradient norm can continue to decrease as $T$ increase, \cite{Jin2020StochasticSignSF} prove that Stochastic-Sign SGD can achieve a convergence rate of $\mathcal{O}\left(d^{3/8}T^{-1/8}\right)$ in terms of the $l_2$-norm, which  is further improved to $\mathcal{O}\left(d^{1/4}T^{-1/4}\right)$ via the variance reduction technique~\citep{NeurIPS:2024:Jiang:B}.

\section{Convergence analysis for the Lion optimizer}
In this section, we first outline the assumptions used in our analysis, followed by the convergence guarantees for the traditional Lion optimizer. To further accelerate the convergence rate, we introduce a variance reduction version of the Lion optimizer and present an improved convergence result.

\subsection{Assumptions}
We begin by listing the assumptions required for our analysis, which are standard in the study of convergence rates in stochastic optimization.
\begin{ass}\label{ass:1} (Smoothness) The objective function $f$ is $L$-smooth, i.e.,  
\begin{align*}
    \Norm{\nabla f(\x) - \nabla f(\y)} \leq L \Norm{\x - \y}.
\end{align*}
\end{ass}

For the analysis of the Lion optimizer with variance reduction, we introduce the following average smoothness assumption, which is also commonly used in the variance reduction literature~\citep{pmlr-v139-li21a,arxiv.1703.00102,Fang2018SPIDERNN,Wang2018SpiderBoostAC,cutkosky2019momentum,levy2021storm}.

\begin{ass}\label{ass:2} (Average smoothness) The stochastic objective function $f(\cdot; \xi)$ is $L$-smooth, satisfying
\begin{align*}
    \mathbb{E}_{\xi}\left[\left\|\nabla f(\x; \xi) - \nabla f(\y; \xi)\right\|^2\right] \leq L^2 \|\x - \y\|^2.
\end{align*}
\end{ass}

\begin{ass}\label{ass:3} The stochastic gradient is unbiased and its noise is bounded. 
\begin{equation*}
\mathbb{E}_{\xi}\left[\nabla f(\x; \xi)\right] = \nabla f(\x), \quad \mathbb{E}_{\xi}\left[\left\|\nabla f(\x; \xi) - \nabla f(\x)\right\|^2\right] \leq \sigma^2.
\end{equation*}
\end{ass}

\begin{ass}\label{ass0} $f_{*}=\inf_{\x} f(\x) \geq-\infty$ and $f\left(\x_{1}\right)-f_{*} \leq \Delta_{f}$ for the initial solution $\x_{1}$.
\end{ass}
\textbf{Remark:} Previous analysis of the Lion optimizer assumes that the objective function $f$ is coercive, such that
\begin{equation*}
 \lim_{\Norm{\x} \to \infty} f(\x) \to \infty,
\end{equation*} 
which is not required in our analysis.

\subsection{Convergence rate for the traditional Lion optimizer}
We first introduce the original Lion optimizer in Algorithm~\ref{alg:lion}~(v1). The Lion optimizer uses the momentum estimators to track the gradient, denoted as $\v_t$ and $\m_t$, and utilizes the sign operation of $\v_t$ for computing the updates. The decoupled weight decay is also incorporated in the update, with $\lambda$ representing the strength of the decay. For the first step ($t=1$), we initialize as $\v_1 = \m_1 = \nabla f(\x_1; \xi_1)$. For the update used in the algorithm, we first have the following lemma.

\begin{algorithm}[t]
    \caption{Lion~(v1: original version, v2: variance reduction version.)}
    \label{alg:lion}
    \begin{algorithmic}[1]
    \STATE {\bfseries Input:} Momentum parameters $\beta_1$, $\beta_2$, learning rate $\eta$, weight decay parameter $\lambda$.
    \FOR{time step $t = 1$ {\bfseries to} $T$}
        \STATE Update $\v_t = (1-\beta_1) \m_{t-1} + \beta_1 \nabla f(\x_t; \xi_t)$
        \STATE (v1) Update $\m_t = (1-\beta_2) \m_{t-1} + \beta_2 \nabla f(\x_t; \xi_t)$
        \STATE (v2) Update $\m_t = (1-\beta_2) \m_{t-1} + \beta_2 \nabla f(\x_t; \xi_t) \textcolor{blue}{ + (1-\beta_2) \left( \nabla f(\x_t; \xi_t) - \nabla f(\x_{t-1}; \xi_t) \right)}$
        \STATE Update $\x_{t+1} = \x_t - \eta \left( \operatorname{sign}(\v_t) + \lambda \x_t \right)$
    \ENDFOR
    \end{algorithmic}
\end{algorithm}
\begin{lem}\label{lem}
For the update $\x_{t+1} = \x_t - \eta \left( \operatorname{sign}(\v_t) + \lambda \x_t \right)$, as long as the initial point satisfies $\Norm{\x_1}_\infty \leq \eta$, we can ensure that each $\x_t$ is bounded such that
\begin{align*}
    \Norm{\x_t}_\infty \leq \eta t , \quad
    \Norm{\x_t}^2 \leq 2\eta^2 t^2 d.
\end{align*}
Additionally, by setting $\lambda \leq \frac{1}{2\eta T}$, we can show that the update step is also bounded, i.e.,
\begin{align*}
    \Norm{\x_{t+1} - \x_t}^2 \leq 4\eta^2 d.
\end{align*}
\end{lem}
Using the above lemma in our theoretical analysis allows us to avoid the coerciveness assumption, which is required in previous literature. Next, we provide the theoretical guarantee for the Lion optimizer.

\begin{theorem}\label{thm1++}
Under Assumptions~\ref{ass:1}, \ref{ass:3} and \ref{ass0}, by setting $\beta_2^2 \leq \beta_1 \leq \sqrt{\beta_2}$, $\beta_2 = \mathcal{O}(T^{-1/2})$, $\eta = \mathcal{O}(d^{-1/2} T^{-3/4})$, $\lambda \leq \frac{1}{2\eta T}$ and $\Norm{\x_1}_\infty \leq \eta$, the Lion optimizer~(v1) ensures that
\begin{align*}
    \frac{1}{T} \sum_{t=1}^T \mathbb{E} \left[\|\nabla f(\x_t)\|_1 \right] \leq \mathcal{O}\left( \frac{d^{1/2}}{T^{1/4}} \right).
\end{align*}
\end{theorem}
\textbf{Remark:} The above result matches the $\mathcal{O}(T^{-1/4})$ lower bound for stochastic optimization under standard smoothness assumption~\citep{Arjevani2019LowerBF}. Compared to previous works, we achieve this rate without the need for additional assumptions, such as the coerciveness of the objective function~\citep{dong2024}.

\subsection{Lion optimizer with variance reduction}
While the $\mathcal{O}(T^{-1/4})$ convergence rate is optimal under the standard smoothness, it is well-known that this rate can be further improved using the average smoothness condition~\citep{Fang2018SPIDERNN,Wang2018SpiderBoostAC,cutkosky2019momentum}. In this subsection, we introduce the Lion optimizer with variance reduction (Lion-VR) to achieve a faster convergence rate.

Note that the traditional Lion optimizer uses two momentum-based estimators $\v_t$ and $\m_t$ to track the gradient. For our Lion-VR method, we modify the $\m_t$ estimator via the variance reduction technique STORM~\citep{cutkosky2019momentum}, such that
\begin{align*}
    \m_t =  (1-\beta_2) \m_{t-1 } + \beta_2 \nabla f(\x_t;\xi_t) \textcolor{blue} {+ (1-\beta_2)\left(\nabla f(\x_t;\xi_t)-\nabla f(\x_{t-1};\xi_t) \right)}.
\end{align*}
The difference lies in the last term, which ensures the estimation error $\mathbb{E} \left[ \Norm{\nabla f(\x_t) - \v_t}^2 \right]$ decays more rapidly over time. This newly proposed algorithm is outlined in Algorithm~\ref{alg:lion}~(v2), named as Lion-VR, with the modification appearing in step 5. For the first time step ($t=1$), we initialize as $\v_1 = \m_1 = \frac{1}{B_0} \sum_{i=1}^{B_0} \nabla f(\x_1; \xi_1^i)$, where $B_0 = \mathcal{O}(T^{1/3})$ is the batch size used in the initial iteration. We can now establish the improved convergence guarantee for this method in the following theorem.

\begin{theorem}\label{thm1+++}
Under Assumptions~\ref{ass:2}, \ref{ass:3} and \ref{ass0}, by setting $\beta_2 \leq \beta_1 \leq \sqrt{\beta_2}$, $\beta_2 = \mathcal{O}(T^{-2/3})$, $\eta = \mathcal{O}(d^{-1/2}T^{-2/3})$, $\lambda \leq \frac{1}{2\eta T}$, and $\Norm{\x_1}_\infty \leq \eta$, the Lion-VR optimizer ensures that
\begin{align*}
    \frac{1}{T} \sum_{t=1}^T \mathbb{E} \left[\|\nabla f(\x_t)\|_1 \right] \leq \mathcal{O}\left( \frac{d^{1/2}}{T^{1/3}} \right).
\end{align*}
\end{theorem}
\textbf{Remark:} The Lion-VR method achieves a better convergence rate of $\mathcal{O}(T^{-1/3})$, which matches the optimal bound for stochastic optimization under the average smoothness assumption~\citep{pmlr-v139-li21a,Arjevani2019LowerBF}.

\section{Convergence analysis for distributed Lion optimizer}
In this section, we investigate the convergence of the Lion optimizer in distributed settings. We first introduce the problem setup and the assumptions used, followed by the distributed Lion optimizer and its variance-reduced version, along with their corresponding convergence guarantees.

\subsection{Problem setup and assumptions}
We begin by considering the following distributed learning problem:
\begin{align}\label{problem3}
    \min_{\x \in \R^d} f(\x) \coloneqq \frac{1}{n}\sum_{j=1}^{n} f_j(\x), \quad f_j(\x) = \mathbb{E}_{\xi^j \sim \mathcal{D}_j} \left[f_j(\x; \xi^j)\right],
\end{align}
where $\mathcal{D}_j$ denotes the data distribution on node $j$, and $f_j(\x)$ is the corresponding loss function. Unlike the homogeneous setting, where the data distribution $\mathcal{D}_j$ and the loss function $f_j$ are assumed to be identical across nodes, we investigate the more challenging heterogeneous setting, where the data and loss function on each node can differ significantly.

Next, we list the assumptions used in our analysis, which are commonly used in stochastic distributed optimization.

\begin{ass}\label{ass:1+} (Smoothness) The objective function on each node is $L$-smooth, such that 
\begin{align*}
    \Norm{\nabla f_j(\x) - \nabla f_j(\y)} \leq L \Norm{\x - \y}.
\end{align*}
\end{ass}

For the variance reduction version of the Lion optimizer, we use the average smoothness assumption instead of the smoothness assumption.
\begin{ass}\label{ass:2+} (Average smoothness) The stochastic objective function on each node is $L$-smooth such that
\begin{align*}
    \mathbb{E}_{\xi}\left[\left\|\nabla f_j(\x; \xi) - \nabla f_j(\y; \xi)\right\|^2\right] \leq L^2 \|\x - \y\|^2.
\end{align*}
\end{ass}

\begin{ass}\label{ass:3+} The stochastic gradient on each node is unbiased, and its noise is bounded, such that 
\begin{equation*}
    \mathbb{E}_{\xi}\left[\nabla f_j(\x; \xi)\right] = \nabla f_j(\x), \quad \mathbb{E}_{\xi}\left[\left\|\nabla f_j(\x; \xi) - \nabla f_j(\x)\right\|^2\right] \leq \sigma^2.
\end{equation*} 
\end{ass}

\begin{ass}\label{ass0+} $f_{*}=\inf_{\x} f(\x) \geq-\infty$ and $f\left(\x_{1}\right)-f_{*} \leq \Delta_{f}$ for the initial solution $\x_{1}$.
\end{ass}

\subsection{Distributed Lion optimizers and convergence guarantees}
In the distributed setting, we first compute the momentum estimators $\v_t^j$ and $\m_t^j$ on each node $j$. For the first step ($t=1$), we initialize $\m_1 = \v_1 = \nabla f(\x_1; \xi_1)$. Then, the parameter server collects each gradient estimator $\v_t^j$ from the nodes and computes $\v_t = \sum_{j=1}^n \v_t^j$. This vector $\operatorname{sign}(\v_t)$ is then sent back to each node, which uses $\operatorname{sign}(\v_t)$ to update the variables. The complete algorithm is outlined in Algorithm~\ref{alg3}~(v1), referred to as the distributed Lion (Dis-Lion) optimizer. Next, we provide the theoretical guarantees for this algorithm.

\begin{algorithm}[!t]
    \caption{Distributed Lion Optimizer}
    \label{alg3}
    \begin{algorithmic}[1]
    \STATE {\bfseries Input:} Momentum parameters $\beta_1$, $\beta_2$, learning rate $\eta$, weight decay parameter $\lambda$.
    \FOR{time step $t = 1$ {\bfseries to} $T$}
        \FOR{node $j = 1$ {\bfseries to} $n$}
            \STATE (v1)  $\v_t^j = (1-\beta_1) \m_{t-1}^j + \beta_1 \nabla f_j(\x_t; \xi_t^j)$
            \STATE \qquad  $\m_t^j = (1-\beta_2) \m_{t-1}^j + \beta_2 \nabla f_j(\x_t;\xi_t^{j}) $
            \STATE (v2) $\v_t^j = (1-\beta_1) \m_{t-1}^j + \beta_1 \nabla f_j(\x_t; \xi_t^j)\textcolor{blue} {+ (1-\beta_1)\left(\nabla f_j(\x_t;\xi_t)-\nabla f_j(\x_{t-1};\xi_t) \right)}$ 
            \STATE \qquad $\m_t^j = (1-\beta_2) \m_{t-1}^j + \beta_2 \nabla f_j(\x_t;\xi_t^{j})  \textcolor{blue} {+ (1-\beta_2)\left(\nabla f_j(\x_t;\xi_t)-\nabla f_j(\x_{t-1};\xi_t) \right)}$ 
            \STATE $\v_t = \sum_{j=1}^n \v_t^j$
            \STATE Update $\x_{t+1} = \x_t - \eta \left( \operatorname{sign}(\v_t) + \lambda \x_t \right)$
        \ENDFOR
    \ENDFOR
    \end{algorithmic}
\end{algorithm}

\begin{theorem}\label{thm3}
Under Assumptions~\ref{ass:1+}, \ref{ass:3+} and \ref{ass0+}, by setting $\beta_2^2 \leq \beta_1 \leq \sqrt{\beta_2}$, $\beta_2 = \mathcal{O}(n^{1/2} T^{-1/2})$, $\eta = \mathcal{O}(n^{1/4} d^{-1/2} T^{-3/4})$, $\lambda \leq \frac{1}{2\eta T}$, $T \ge n$, and $\Norm{\x_1}_\infty \leq \eta$, the distributed Lion optimizer (v1) ensures that
\begin{align*}
    \frac{1}{T} \sum_{t=1}^T \mathbb{E} \left[ \|\nabla f(\x_t)\|_1 \right] \leq \mathcal{O}\left( \frac{d^{1/2}}{n^{1/4} T^{1/4}} \right).
\end{align*}
\end{theorem}
\textbf{Remark:} The proposed method achieves the optimal $\mathcal{O}(T^{-1/4})$ convergence under the standard smoothness assumption, matching the results obtained in the centralized settings.

To improve the convergence rate using the average smoothness assumption, we further introduce the variance reduction technique into the momentum estimator $\m_t$ and $\v_t$. Similar to the centralized setting, we apply the STORM estimator on each node $j$ such that
\begin{align*}
    \m_t^j = (1-\beta_2) \m_{t-1}^j + \beta_2 \nabla f_j(\x_t; \xi_t^j) + (1-\beta_2) \left( \nabla f_j(\x_t; \xi_t) - \nabla f_j(\x_{t-1}; \xi_t) \right),\\
    \v_t^j = (1-\beta_2) \m_{t-1}^j + \beta_2 \nabla f_j(\x_t; \xi_t^j) + (1-\beta_2) \left( \nabla f_j(\x_t; \xi_t) - \nabla f_j(\x_{t-1}; \xi_t) \right).
\end{align*}

The resulting algorithm is outlined in Algorithm~\ref{alg3}~(v2), referred to as Dis-Lion-VR, with the modification appearing in step 6. For the first step ($t=1$), we initialize $\v_1 = \m_1 = \frac{1}{B_0} \sum_{i=1}^{B_0} \nabla f(\x_1; \xi_1^i)$, where $B_0 = \mathcal{O}(n^{-2/3}T^{1/3})$ is the batch size used in the first iteration.

\begin{theorem}\label{thm4}
Under Assumptions~\ref{ass:2+}, \ref{ass:3+} and \ref{ass0+}, by setting that $\beta_1 \leq \sqrt{\beta_2}$, $\beta_2 = \mathcal{O}(n^{1/3} T^{-2/3})$, $\eta = \mathcal{O}(n^{1/3} d^{-1/2} T^{-2/3})$, $\lambda \leq \frac{1}{2\eta T}$, $T \geq n^2$, and $\Norm{\x_1}_\infty \leq \eta$, the distributed Lion optimizer (v2) ensures that
\begin{align*}
    \frac{1}{T} \sum_{t=1}^T \mathbb{E} \left[ \|\nabla f(\x_t)\|_1 \right] \leq \mathcal{O}\left( \frac{d^{1/2}}{n^{1/3} T^{1/3}} \right).
\end{align*}
\end{theorem}
\textbf{Remark:} The above theorem establishes the $\mathcal{O}(T^{-1/3})$ convergence rate under the average smoothness assumption, matching the results in the centralized setting and the corresponding lower bound~\citep{pmlr-v139-li21a,Arjevani2019LowerBF}.

\section{Communication-efficient distributed Lion optimizer}
In the previous section, we note that the Distributed Lion optimizer is communication-efficient since the information sent back to each node is the 1-bit sign information~(e.g., $\operatorname{sign}(\v_t)$). However, during the update, each node must send the gradient estimator $\v_t^j$ to the parameter server, which is no longer 1-bit sign information. A natural question is whether each node can send only the sign information of $\v_t^j$ to the server, thereby making the algorithm communication-efficient in both directions. In this section, we investigate the distributed Lion optimizer with 1-bit sign compression in both directions.

The most straightforward approach is to apply the sign operation twice, leading to the following update rule:
\begin{align*}
    \x_{t+1} = \x_t - \eta \left( \operatorname{sign}\left( \frac{1}{n} \sum_{j=1}^n \operatorname{sign}(\v_t^j) \right) + \lambda \x_t \right).
\end{align*}
However, the sign operation is a biased estimator, and applying it twice may introduce significant estimation error. To mitigate this, we use the following unbiased sign operation, which is widely adopted in previous sign-based optimization algorithms.
\begin{defi}
    For any vector \( \v \) with \( \|\v\|_{\infty} \leq R \), the unbiased sign operation $\operatorname{S_\textit{R}}(\v)$ is defined as:
    \begin{align}\label{mapping}
    [\operatorname{S_\textit{R}}(\v)]_k = \begin{cases}
        1, & \text{with probability } \frac{R + [\v]_k}{2R}, \\
        -1, & \text{with probability } \frac{R - [\v]_k}{2R}.
    \end{cases}
    \end{align}
\end{defi}
\textbf{Remark:} This operation provides an unbiased estimation of \( \frac{\v}{R} \), such that \( \mathbb{E}[\operatorname{S_\textit{R}}(\v)] = \frac{\v}{R} \).

Note that the unbiased sign operation requires the input vector to be bounded. Therefore, we assume the following bounded gradients condition, which is also used in previous sign-based distributed algorithms~\citep{Jin2020StochasticSignSF, pmlr-v202-sun23l}.

\begin{ass}(Bounded Gradients)\label{bg1}
    For each node $j \in [n]$, we assume that the gradient of the loss function is bounded, i.e., 
    \[
    \sup_{\x} \|\nabla f_j(\x; \xi)\|_\infty \leq G.
    \]
\end{ass}
Next, we introduce the proposed algorithm. Most steps are similar to those in Algorithm~\ref{alg3}, with the key difference being in the parameter update, where we now use:
\begin{align*}
    \x_{t+1} = \x_t - \eta \left(\textcolor{blue} {Q_2}\left(\frac{1}{n} \sum_{j=1}^n \textcolor{blue} {Q_1}(\v_t^j)\right) +\lambda \x_t \right),
\end{align*}
where \( Q_1 \) and \( Q_2 \) represent the sign or unbiased sign operation, which may vary depending on the specific theorem. The complete algorithm is described in Algorithm~\ref{alg4}~(v1). For the first iteration ($t=1$), we initialize $\v_1^j = \m_1^j = \nabla f_j(\x_1; \xi_1^j)$.

\begin{algorithm}[!t]
    \caption{Communication-efficient Distributed Lion Optimizer}
    \label{alg4}
    \begin{algorithmic}[1]
    \STATE {\bfseries Input:} Momentum parameters $\beta_1$, $\beta_2$, learning rate $\eta$, weight decay parameter $\lambda$.
    \FOR{time step $t = 1$ {\bfseries to} $T$}
        \FOR{node $j = 1$ {\bfseries to} $n$}
            \STATE Compute $\v_t^j = (1-\beta_1) \m_{t-1}^j + \beta_1 \nabla f_j(\x_t; \xi_t^j)$
            \STATE (v1) $\m_t^j = (1-\beta_2) \m_{t-1}^j + \beta_2 \nabla f_j(\x_t;\xi_t^{j}) $
        \STATE (v2) $\m_t^j = (1-\beta_2) \m_{t-1}^j + \beta_2 \nabla f_j(\x_t;\xi_t^{j})  \textcolor{blue} {+ (1-\beta_2)\left(\nabla f_j(\x_t;\xi_t)-\nabla f_j(\x_{t-1};\xi_t) \right)}$ 
            \STATE $\v_t = \frac{1}{n} \sum_{j=1}^n Q_1(\v_t^j)$
            \STATE Update $\x_{t+1} = \x_t - \eta \left( Q_2(\v_t) + \lambda \x_t \right)$
        \ENDFOR
    \ENDFOR
    \end{algorithmic}
\end{algorithm}

Now, we present the convergence guarantees for the proposed algorithm. In the first version, we use the traditional sign operation on the server, i.e., $Q_2(\cdot) = \operatorname{sign(\cdot)}$, and employ the unbiased sign function on each node, such that $Q_1(\cdot) = \text{S}_G(\cdot)$. In this case, we can establish the following convergence results.

\begin{theorem}\label{thm5}
Under Assumptions~\ref{ass:1+}, \ref{ass:3+}, \ref{ass0+} and \ref{bg1}, set that  $\beta_2^2 \leq \beta_1 \leq \sqrt{\beta_2}$, $\beta_2 = \frac{1}{2}$, $\lambda \leq \frac{1}{2\eta T}$, $Q_1(\cdot) = \text{S}_G(\cdot)$, and $Q_2(\cdot) = \operatorname{sign}(\cdot)$. Then, we have the following convergence guarantees:

\begin{enumerate}
    \item By setting the learning rate $\eta = \mathcal{O}\left( \frac{1}{T^{1/2} d^{1/2}} \right)$, Algorithm~\ref{alg4}~(v1) ensures:
    \begin{align*}
        \frac{1}{T} \sum_{t=1}^T \mathbb{E} \left[\|\nabla f(\x_t)\|_1 \right] \leq \mathcal{O}\left( \frac{d^{1/2}}{T^{1/2}} + \frac{d}{n^{1/2}} \right).
    \end{align*}
    
    \item By setting the learning rate $\eta = \mathcal{O}\left( \frac{1}{n^{1/2}} \right)$, Algorithm~\ref{alg4}~(v1) ensures:
    \begin{align*}
        \frac{1}{T} \sum_{t=1}^T \mathbb{E} \left[\|\nabla f(\x_t)\|_1 \right] \leq \mathcal{O}\left( \frac{n^{1/2}}{T} + \frac{d}{n^{1/2}} \right).
    \end{align*}
\end{enumerate}
\end{theorem}
\textbf{Remark:} By adjusting the learning rate, we obtain the two convergence guarantees above. The second one is preferable when $n \leq T d$. This theorem shows that the gradient $l_1$-norm can quickly converge to $\mathcal{O}\left(\frac{d}{n^{1/2}}\right)$.

In many practical scenarios, we want the gradient norm to decrease continuously. To achieve this, we further apply an unbiased sign operation on the server, setting $Q_2(\cdot) = \text{S}_1(\cdot)$. Note that the $S_1(\cdot)$ operation is valid because the input is the average of the sign information $\frac{1}{n} \sum_{j=1}^n {Q_1}(\v_t^j)$, which lies within $[-1,1]$. With this modification, we obtain the following convergence guarantee.

\begin{theorem}\label{thm7}
Under Assumptions~\ref{ass:1+}, \ref{ass:3+}, \ref{ass0+} and \ref{bg1}, by setting $\eta = \mathcal{O}\left( \min \left\{ \frac{1}{T^{1/2} d^{1/2}}, \frac{n^{2/5}}{T^{3/5} d^{1/5}} \right\} \right)$, $\beta_2^2 \leq \beta_1 \leq \sqrt{\beta_2}$, $\beta_2 = \mathcal{O}\left( n^{1/3} \eta^{2/3} d^{1/3} \right)$, $\lambda \leq \min\left\{\frac{\sqrt{L}}{T\sqrt{\eta G}},\frac{1}{2\eta T} \right\}$, $Q_1(\cdot) = \text{S}_G(\cdot)$, $Q_2(\cdot) = \text{S}_1(\cdot)$, $T \ge n$, and $\Norm{\x_1}_\infty \leq \eta$, Algorithm~\ref{alg4}~(v1) ensures:
\begin{align*}
    \frac{1}{T} \sum_{t=1}^T \mathbb{E} \left[\|\nabla f(\x_t)\|_1 \right] \leq \mathcal{O}\left( \max \left\{ \frac{d^{1/4}}{T^{1/4}}, \frac{d^{1/10}}{n^{1/5}T^{1/5}} \right\} \right).
\end{align*}
\end{theorem}
\textbf{Remark:} The above rate can converge to zero as the iteration number $T \to \infty$, ensuring that the gradient norm becomes smaller as $T$ increases.

Next, we apply the variance reduction technique along with the average smoothness assumption to further improve the convergence rate. The only modification is that we use the STORM estimator $\m_t^i$ in step 6. The updated algorithm is described in Algorithm~\ref{alg4}~(v2). We now present the convergence guarantee for this improved method.

\begin{theorem}\label{thm8}
Under Assumptions~\ref{ass:2+}, \ref{ass:3+}, \ref{ass0+} and \ref{bg1}, by setting that $\beta_1 \leq \sqrt{\beta_2}$, $\beta_2 = \mathcal{O}\left( \frac{1}{T^{1/2}} \right)$, $\eta = \mathcal{O}\left( \frac{1}{d^{1/2} T^{1/2}} \right)$, $\lambda \leq \min\left\{\frac{\sqrt{L}}{T\sqrt{\eta G}},\frac{1}{2\eta T} \right\}$, $Q_1(\cdot) = \text{S}_G(\cdot)$, $Q_2(\cdot) = \text{S}_1(\cdot)$, and $\Norm{\x_1}_\infty \leq \eta$, Algorithm~\ref{alg4}~(v2) ensures:
\begin{align*}
    \frac{1}{T} \sum_{t=1}^T \mathbb{E} \left[\|\nabla f(\x_t)\|_1 \right] \leq \mathcal{O}\left( \frac{d^{1/4}}{T^{1/4}} \right).
\end{align*}
\end{theorem}
\textbf{Remark:} This rate can also approach zero as the iteration number increases, and the obtained convergence rate is better than the one derived in Theorem~\ref{thm7}.

\section{Conclusion}
This paper investigates the convergence guarantees of the Lion optimizer in both centralized and distributed settings. We first demonstrate that the Lion optimizer and its variance-reduced version achieve convergence rates of $\mathcal{O}(d^{1/2}T^{-1/4})$ and $\mathcal{O}(d^{1/2}T^{-1/3})$, respectively. We then extend these results to their distributed versions, showing that similar convergence guarantees hold in distributed settings. To further improve communication efficiency, we develop a distributed version of the Lion optimizer with sign compression in both directions, yielding convergence rates of $\mathcal{O}\left( \frac{d^{1/2}}{T^{1/2}} + \frac{d}{n^{1/2}} \right)$ and $\mathcal{O}\left( \frac{n^{1/2}}{T} + \frac{d}{n^{1/2}} \right)$. Finally, by employing the unbiased sign operation in both directions, we achieve  convergence rates of $\mathcal{O}\left( \max \left\{ \frac{d^{1/4}}{T^{1/4}}, \frac{d^{1/10}}{n^{1/5}T^{1/5}} \right\} \right)$ and $\mathcal{O}\left( \frac{d^{1/4}}{T^{1/4}} \right)$, respectively. 

\vskip 0.2in
\bibliography{ref}

\newpage
\appendix
\section{Proof of Lemma~\ref{lem}}
For the update step $\x_{t+1} = \x_t - \eta \left( \operatorname{sign}(\v_t) + \lambda \x_t \right)$, as long as the initial point satisfies the condition $\Norm{\x_1}_\infty \leq \eta$, we can ensure that each $\x_t$ is bounded such that
\begin{align*}
    &\Norm{\x_{t}}_\infty \\
    =& \Norm{(1-\eta\lambda)\x_{t-1} - \eta    \operatorname{sign}\left(\v_{t-1}\right) }_\infty \\
    \leq&  (1-\eta\lambda)\Norm{\x_{t-1}}_\infty + \eta    \Norm{\operatorname{sign}\left(\v_{t-1}\right) }_\infty \\
     \leq& (1-\eta\lambda)^{t-1}\Norm{\x_1}_\infty +\eta (t-1) \max_{i \in \{1,\cdots,t-1\}}\left\{\Norm{\operatorname{sign}\left(\v_i \right)}_\infty\right\} \\
     \leq& \Norm{\x_1}_\infty +\eta (t-1) \\
     \leq &\eta t.
\end{align*}
Also, we can ensure that
\begin{align*}
    &\Norm{\x_{t}} \\
    =& \Norm{(1-\eta\lambda)\x_{t-1} - \eta    \operatorname{sign}\left(\v_{t-1}\right) } \\
    \leq &(1-\eta\lambda)\Norm{\x_{t-1}} + \eta    \Norm{\operatorname{sign}\left(\v_{t-1}\right) } \\
    \leq &(1-\eta\lambda)\Norm{\x_{t-1}} + \eta \sqrt{d} \\
    \leq & (1-\eta\lambda)^{t-1} \Norm{\x_1} + \eta(t-1) \sqrt{d},
\end{align*}
as well as
\begin{align*}
    &\Norm{\x_{t}}^2 \\
    \leq & 2(1-\eta\lambda)^{2(t-1)} \Norm{\x_1}^2 + 2\eta^2(t-1)^2 d \\
    \leq & 2\Norm{\x_1}^2 +2\eta^2 (t-1)^2 d \\
    \leq & 2\Norm{\x_1}_\infty^2 d +2\eta^2 (t-1)^2 d \\
    \leq& 2\eta^2 t^2 d.
\end{align*}
Next, we show that the decision variable update is bounded by setting $\lambda \leq \frac{1}{2\eta T}$, such that
\begin{align*}
    &\Norm{\x_{t+1}-\x_{t}}^2 \\
    = &\Norm{\eta\lambda\x_{t} + \eta    \operatorname{sign}\left(\v_{t}\right) }^2 \\
    \leq & 2\eta^2\lambda^2\Norm{\x_{t}}^2 + 2\eta^2    \Norm{\operatorname{sign}\left(\v_{t}\right) }^2 \\
    \leq & 2\eta^2\lambda^2 \Norm{\x_t}^2 +2\eta^2 d \\
    \leq & 2\eta^2\lambda^2 \cdot 2\eta^2t^2d +2\eta^2 d \\
    \leq & 2\eta^2 d(2\eta^2\lambda^2 T^2+1) \\
    \leq & 4\eta^2 d,
\end{align*}
which finishes the proof of Lemma~\ref{lem}.

\section{Proof of Theorem~\ref{thm1++}}
Due to the smoothness assumption~(Assumption~\ref{ass:1}), we have that
\begin{equation}\label{L-smooth}
\begin{aligned}
f(\x_{t+1}) &\leq f(\x_t) + \left\langle \nabla f(\x_t), \x_{t+1} - \x_t \right\rangle + \frac{L}{2} \| \x_{t+1} - \x_t \|^2 \\
&= f(\x_t) -\eta \left\langle \nabla f(\x_t),  \sign(\v_t) \right\rangle -\eta \lambda \left\langle \nabla f(\x_t),  \x_t \right\rangle+ 2\eta^2 Ld \\
&\leq f(\x_t) + \eta\left\langle \nabla f(\x_t),  \sign(\nabla f(\x_t)) -  \sign(\v_t) \right\rangle - \eta \left\langle \nabla f(\x_t), \sign(\nabla f(\x_t)) \right\rangle \\
&\quad -\eta \lambda \left\langle \nabla f(\x_t),  \x_t \right\rangle+ 2\eta^2 Ld\\
&= f(\x_t) +\eta\left\langle \nabla f(\x_t),  \sign(\nabla f(\x_t)) -  \sign(\v_t) \right\rangle  - \eta \| \nabla f(\x_t) \|_1 \\
&\quad +\eta \lambda \Norm{\nabla f(\x_t)}_1\Norm{\x_t}_\infty + 2\eta^2 Ld\\
&= f(\x_t) +\eta \sum_{i=1}^{d} \left\langle [\nabla f(\x_t)]_i, \sign([\nabla f(\x_t)]_i) - \sign([\v_t]_i) \right\rangle  - \frac{\eta}{2} \| \nabla f(\x_t) \|_1 + 2\eta^2 Ld\\
&\leq f(\x_t) +\eta\sum_{i=1}^{d} 2\left|[\nabla f(\x_t)]_i\right| \cdot \mathbb{I} \left( \sign(\left[\nabla f(\x_t)\right]_i) \neq \sign([\v_t]_i) \right) \\
&\quad -  \frac{\eta}{2} \| \nabla f(\x_t) \|_1 + 2\eta^2 Ld\\
&\leq f(\x_t) +\eta\sum_{i=1}^{d} 2|[\nabla f(\x_t)]_i - [\v_t]_i| \cdot \mathbb{I} \left( \sign(\left[\nabla f(\x_t)\right]_i) \neq \sign([\v_t]_i) \right) \\
&\quad-  \frac{\eta}{2} \| \nabla f(\x_t) \|_1 + 2\eta^2 Ld\\
&\leq f(\x_t) +\eta\sum_{i=1}^{d} 2|[\nabla f(\x_t)]_i - [\v_t]_i| - \frac{\eta}{2} \| \nabla f(\x_t) \|_1 + 2\eta^2 Ld\\
&= f(\x_t) +2\eta\Norm{\nabla f(\x_t) - \v_t}_1  - \frac{\eta}{2} \| \nabla f(\x_t) \|_1 + 2\eta^2 Ld\\
&\leq f(\x_t) +2\eta\sqrt{d} \Norm{\nabla f(\x_t) - \v_t}  - \frac{\eta}{2} \| \nabla f(\x_t) \|_1 + 2\eta^2 Ld.
\end{aligned}
\end{equation}
Summing up and rearranging the equation~(\ref{L-smooth}), we derive:
\begin{equation}\label{smooth}
    \begin{split}
        &\E\left[\frac{1}{T}\sum_{t=1}^{T} \|\nabla f(\x_t)\|_1\right]\\ 
        \leq &\frac{2\left(f(\x_1) - f(\x_{T+1})\right)}{\eta T} +  4\sqrt{d} \cdot \E\left[\frac{1}{T}\sum_{t=1}^{T} \|\nabla f(\x_t) - \v_t\|\right] + 4\eta L d \\
        \leq & \frac{2\Delta_f}{\eta T} +4\sqrt{d} \cdot \sqrt{\E\left[\frac{1}{T}\sum_{t=1}^{T} \|\nabla f(\x_t) - \v_t\|^2\right]}+ {4\eta L d}.
    \end{split}
\end{equation}
where  $f\left(\x_{1}\right)-f_{*}\leq  \Delta_{f}$, and the second inequality is due to Jensen's Inequality.

Next, we can bound the term $\E\left[\frac{1}{T}\sum_{t=1}^{T} \|\nabla f(\x_t) - \v_t\|^2\right]$ as follows.
\begin{equation}\label{cont}
\begin{split}
        &\E\left[\Norm{\nabla f(\x_{t+1}) -\v_{t+1}}^2\right] \\
        =& \E\left[\Norm{(1-\beta_1)\m_{t} + \beta_1 \nabla f(\x_{t+1};\xi_{t+1})  - \nabla f(\x_{t+1})}^2\right]\\
         = &\E\left[\left\|(1-\beta_1)(\m_{t} - \nabla f(\x_{t})) + \beta_1 \left(\nabla f(\x_{t+1};\xi_{t+1}) - \nabla f(\x_{t+1})  \right)  \right.\right. \\
        & \quad \left.\left.  + (1-\beta_1)\left(\nabla f(\x_{t})-\nabla f(\x_{t+1}) \right)\right\|^2 \right]\\
         =& (1-\beta_1)^2 \E\left[\left\|\m_{t} - \nabla f(\x_{t})+\nabla f(\x_{t}) - \nabla f(\x_{t+1})\right\|^2\right] \\
        & \quad + \beta_1^2 \E\left[\left\|\nabla f(\x_{t+1};\xi_{t+1}) - \nabla f(\x_{t+1})\right\|^2\right] \\
         \leq& (1-\beta_1)^2 (1+\beta_1)\E\left[\left\|\m_{t} - \nabla f(\x_{t}))\right\|^2\right] \\
        &\quad +(1-\beta_1)^2(1+\frac{1}{\beta_1}) \E\left[\left\|\nabla f(\x_{t}) - \nabla f(\x_{t+1})\right\|^2\right] + \beta_1^2 \sigma^2 \\
         \leq & (1-\beta_1)\E\left[\|\m_{t} - \nabla f(\x_{t})\|^2\right]   + \frac{2L^2}{\beta_1} \|\x_{t+1} - \x_{t} \|^2+ \beta_1^2\sigma^2 \\
         \leq & (1-\beta_1)\E\left[\|\m_{t} - \nabla f(\x_{t})\|^2\right]   + \frac{8\eta^2L^2d}{\beta_1}+ \beta_1^2\sigma^2,
    \end{split}
\end{equation}
where the third equality is due to the fact $\E\left[\nabla f(\x_{t+1};\xi_{t+1}) - \nabla f(\x_{t+1}) \right]=0$.

Very similarly, we can obtain that 
\begin{align*}
        &\E\left[\Norm{\nabla f(\x_{t+1}) -\m_{t+1}}^2\right] \\
        =& \E\left[\Norm{(1-\beta_2)\m_{t} + \beta_2 \nabla f(\x_{t+1};\xi_{t+1})  - \nabla f(\x_{t+1})}^2\right]\\
        \leq &  (1-\beta_2)\E\left[\|\m_{t} - \nabla f(\x_{t})\|^2\right]   + \frac{8\eta^2L^2d}{\beta_2}+ \beta_2^2\sigma^2.
\end{align*}
Summing up, we can ensure
\begin{equation}
    \begin{aligned}
     \E\left[\frac{1}{T}\sum_{t=1}^T \|\m_t - \nabla f(\x_t)\|^2\right] &\leq \frac{\E\left[\Norm{\m_1-\nabla f(\x_1)}^2\right]}{\beta_2 T}  + \frac{8\eta^2 L^2  d}{\beta_2^2}+ \beta_2\sigma^2\\
     &\leq \frac{\sigma^2}{\beta_2 T}  + \frac{8\eta^2 L^2  d}{\beta_2^2}+ \beta_2\sigma^2.
\end{aligned}
\end{equation}
That is to say, by setting that $\beta_2^2 \leq \beta_1 \leq \sqrt{\beta_2}$, we can ensure
\begin{equation*}
    \begin{aligned}
     \E\left[\frac{1}{T}\sum_{t=1}^T \|\v_t - \nabla f(\x_t)\|^2\right] &\leq \frac{\sigma^2}{T}+(1-\beta_1)\E\left[\frac{1}{T}\sum_{t=1}^{T-1} \|\m_t - \nabla f(\x_t)\|^2\right]  + \frac{8\eta^2 L^2  d}{\beta_1}+ \beta_1^2\sigma^2\\
     &\leq \frac{\sigma^2}{T}+\frac{\sigma^2}{\beta_2 T}  + \frac{8\eta^2 L^2  d}{\beta_2^2}+ \beta_2\sigma^2   + \frac{8\eta^2 L^2  d}{\beta_1}+ \beta_1^2\sigma^2\\
     &\leq \frac{2\sigma^2}{\beta_2 T} + \frac{16\eta^2 L^2 d}{\beta_2^2} + 2\beta_2 \sigma^2.
\end{aligned}
\end{equation*}
By setting that $\beta_2 = \mathcal{O}\left( T^{-1/2}\right)$, $\eta = \mathcal{O}\left( d^{-1/2}T^{-3/4}\right)$, we observe:
\begin{equation*}
    \begin{split}
        &\E\left[\frac{1}{T}\sum_{t=1}^{T} \|\nabla f(\x_t)\|_1\right] \\
\leq & \frac{2\Delta_f}{\eta T} +4\sqrt{d} \cdot \sqrt{\E\left[\frac{1}{T}\sum_{t=1}^{T} \|\nabla f(\x_t) - \v_t\|^2\right]}+ 4\eta L d \\
\leq & \frac{2\Delta_f}{\eta T} +4\sqrt{d} \cdot \sqrt{\frac{2\sigma^2}{\beta_2 T}  + \frac{16\eta^2 L^2  d}{\beta_2^2}+ 2\beta_2\sigma^2}+ 4\eta L d \\
= &\mathcal{O}\left( \frac{ \left(1+\Delta_f+\sigma+L \right) d^{1/2}}{T^{1/4}}\right) \\
= &\mathcal{O}\left( \frac{d^{1/2}}{T^{1/4}}\right),
    \end{split}
\end{equation*}
which finishes the proof.

\section{Proof of Theorem~\ref{thm1+++}}
Since the variance reduction version only differs from the original version in step 5, the analysis of Theorem~\ref{thm1++} is valid until equation~(\ref{cont}).

The main difference is that we can bound the estimator error $\E\left[\frac{1}{T}\sum_{t=1}^{T} \|\nabla f(\x_t) - \m_t\|^2\right]$ term as follows.
\begin{align*}
        &\quad\ \E\left[\Norm{\nabla f(\x_{t+1}) -\m_{t+1}}^2\right]\\
        & = \E\left[\Norm{(1-\beta_2)\m_{t} + \beta_2 \nabla f(\x_{t+1};\xi_{t+1}) +  (1-\beta_2)(\nabla f(\x_{t+1};\xi_{t+1})-\nabla f(\x_{t};\xi_{t+1})) - \nabla f(\x_{t+1})}^2\right]\\
        & = \E\left[\left\|(1-\beta_2)(\m_{t} - \nabla f(\x_{t})) + \beta_2 \left(\nabla f(\x_{t};\xi_{t+1}) - \nabla f(\x_{t})  \right)  \right.\right. \\
        & \qquad \left.\left.  + \left(\nabla f(\x_{t})-\nabla f(\x_{t+1}) + \nabla f(\x_{t+1};\xi_{t+1}) - \nabla f(\x_{t};\xi_{t+1}) \right)\right\|^2 \right]\\
        & \leq (1-\beta_2)^2 \E\left[\left\|\m_{t} - \nabla f(\x_{t})\right\|^2\right] + 2\beta_2^2 \E\left[\left\|\nabla f(\x_{t};\xi_{t+1}) - \nabla f(\x_{t})\right\|^2\right] \\
        & \qquad +2 \E\left[\left\|\left(\nabla f(\x_{t+1};\xi_{t+1}) - \nabla f(\x_{t};\xi_{t+1})\right) \right\|^2\right]\\
        & \leq  (1-\beta_2)\E\left[\|\m_{t} - \nabla f(\x_{t})\|^2\right]  + {2\beta_2^2\sigma^2} + {2L^2 \|\x_{t+1} - \x_{t} \|^2}\\
        & \leq (1-\beta_2)\E\left[\|\m_{t} - \nabla f(\x_{t})\|^2\right]  + {2\beta_2^2\sigma^2} + {8L^2 \eta^2 d},
\end{align*}
where the first inequality is due to the fact that $(a+b)^2 \leq 2a^2+2b^2$ as well as the fact that $\E\left[\left(\beta \left(\nabla f(\x_{t};\xi_{t+1}\right) - \nabla f(\x_{t})  \right)  +\nabla f(\x_{t})-\nabla f(\x_{t+1}) + \nabla f(\x_{t+1};\xi_{t+1}) - \nabla f(\x_{t};\xi_{t+1}^k)  \right]=0$.

Summing up and noticing that we use a batch size of $B_0$ in the first iteration, we can ensure that
\begin{equation}\label{red}
    \begin{aligned}
     \E\left[\frac{1}{T}\sum_{t=1}^T \|\m_t - \nabla f(\x_t)\|^2\right] \leq& \frac{\E\left[\Norm{\m_1-\nabla f(\x_1)}^2\right]}{\beta_2 T} + {2\sigma^2 \beta_2} + \frac{8L^2 \eta^2 d}{\beta_2}\\
     \leq& \frac{\sigma^2}{B_0\beta_2 T} + {2\sigma^2 \beta_2} + \frac{8L^2 \eta^2 d}{\beta_2}
\end{aligned}
\end{equation}
That is to say, by setting that $\beta_2 \leq \beta_1 \leq \sqrt{\beta_2}$ and $B_0 \beta_2 \leq 1$, we can ensure
\begin{equation*}
    \begin{aligned}
     &\E\left[\frac{1}{T}\sum_{t=1}^T \|\v_t - \nabla f(\x_t)\|^2\right]\\ \leq &\frac{\sigma^2}{T}+(1-\beta_1)\E\left[\frac{1}{T}\sum_{t=1}^{T-1} \|\m_t - \nabla f(\x_t)\|^2\right]  + \frac{8\eta^2 L^2  d}{\beta_1}+ \beta_1^2\sigma^2\\
     \leq &\frac{\sigma^2}{T}+\frac{\sigma^2}{B_0 \beta_2 T} + 2\beta_2\sigma^2  + \frac{8\eta^2 L^2  d}{\beta_2}  + \frac{8\eta^2 L^2  d}{\beta_1}+ \beta_1^2\sigma^2\\
     \leq &\frac{2\sigma^2}{B_0 \beta_2 T} + \frac{16\eta^2 L^2 d}{\beta_2} + 3\beta_2 \sigma^2.
\end{aligned}
\end{equation*}
By setting that $\beta_2 = \mathcal{O}\left( T^{-2/3}\right)$, $\eta = \mathcal{O}\left( d^{-1/2}T^{-2/3}\right)$, $B_0 = \mathcal{O}\left( T^{1/3}\right)$ we observe:
\begin{equation*}
    \begin{split}
        \E\left[\frac{1}{T}\sum_{t=1}^{T} \|\nabla f(\x_t)\|_1\right] 
& \leq \frac{2\Delta_f}{\eta T} +4\sqrt{d} \cdot \sqrt{\E\left[\frac{1}{T}\sum_{t=1}^{T} \|\nabla f(\x_t) - \v_t\|^2\right]}+ 4\eta L d \\
& \leq \frac{2\Delta_f}{\eta T} +4\sqrt{d} \cdot \sqrt{\frac{2\sigma^2}{B_0\beta_2 T}  + \frac{16\eta^2 L^2  d}{\beta_2}+ 3\beta_2\sigma^2}+ 4\eta L d \\
&= \mathcal{O}\left( \frac{ \left(1+\Delta_f+\sigma+L \right) d^{1/2}}{T^{1/3}}\right) \\
&= \mathcal{O}\left( \frac{d^{1/2}}{T^{1/3}}\right),
    \end{split}
\end{equation*}
which finishes the proof.

\section{Proof of Theorem~\ref{thm3}}
Note that the analysis of Theorem~\ref{thm1++} is valid for Theorem~\ref{thm3} until the equation~(\ref{smooth}). That is to say,  we can still ensure
\begin{equation*}
    \begin{split}
        \E\left[\frac{1}{T}\sum_{t=1}^{T} \|\nabla f(\x_t)\|_1\right] 
& \leq \frac{2 \Delta_f}{\eta T} +4\sqrt{d} \cdot \sqrt{\E\left[\frac{1}{T}\sum_{t=1}^{T} \|\nabla f(\x_t) - \v_t\|^2\right]}+ 4\eta L d.
    \end{split}
\end{equation*}
Next, we can bound $\E\left[\frac{1}{T}\sum_{t=1}^{T} \|\nabla f(\x_t) - \v_t\|^2\right] = \E\left[\frac{1}{T}\sum_{t=1}^T \Norm{\frac{1}{n}\sum_{j=1}^n \v_{t}^j - \nabla f(\x_{t})}^2\right]$ as follows.
For each worker $j$, we have the following according to the definition of $\v_t^j$:
\begin{align*}
    \v_{t+1}^j - \nabla f_j(\x_{t+1})
     = &(1-\beta_1)\left(\m_t^j - \nabla f_j(\x_t)\right) + \beta_1 \left(\nabla f_j(\x_{t+1};\xi_{t+1}^{j}) - \nabla f_j(\x_{t+1})\right)\\
    & + (1-\beta_1)  \left( \nabla f_j(\x_{t}) - \nabla f_j(\x_{t+1}) \right).
\end{align*}
Averaging over $\{n\}$ and noting that $\nabla f(\x) = \frac{1}{n}\sum_{j=1}^n \nabla f_j(\x)$, we can obtain:
\begin{align*}
    &\frac{1}{n}\sum_{j=1}^n \v_{t+1}^j - \nabla f(\x_{t+1}) = \frac{1}{n}\sum_{j=1}^n \left(\v_{t+1}^j - \nabla f_j(\x_{t+1})\right)\\
     = &(1-\beta_1)\frac{1}{n}\sum_{j=1}^n \left(\m_t^j - \nabla f_j(\x_t)\right) + \beta_1 \frac{1}{n}\sum_{j=1}^n \left(\nabla f_j(\x_{t+1};\xi_{t+1}^{j}) - \nabla f_j(\x_{t+1})\right)\\
    & \quad + (1-\beta_1) \frac{1}{n}\sum_{j=1}^n \left(\nabla f_j(\x_{t}) - \nabla f_j(\x_{t+1}) \right).
\end{align*}
Then we have
\begin{align}
\begin{split}\label{SSR}
     & \E\left[\left\|\frac{1}{n}\sum_{j=1}^n \v_{t+1}^j - \nabla f(\x_{t+1})\right\|^2\right]\\
     \leq &(1-\beta_1)\E\left[\Norm{ \frac{1}{n}\sum_{j=1}^n\left(\m_{t}^j - \nabla f_j(\x_{t})\right)}^2\right]+ \beta_1^2 \frac{1}{n^2}\sum_{j=1}^n\E\left[\Norm{ \nabla f_j(\x_{t+1};\xi_{t+1}^{j}) - \nabla f_j(\x_{t+1})}^2\right]\\
     & +\frac{2}{\beta_1 n}\sum_{j=1}^n\E\left[\Norm{ \nabla f_j(\x_{t+1}) - \nabla f_j(\x_{t})}^2\right]\\
     \leq &(1-\beta_1)\E\left[\Norm{\frac{1}{n}\sum_{j=1}^n \left(\m_{t}^j - \nabla f_j(\x_{t})\right)}^2\right] + \frac{\beta_1^2\sigma^2}{n} + \frac{2L^2}{\beta_1}\Norm{\x_{t+1}-\x_t}^2\\
     \leq &(1-\beta_1)\E\left[\Norm{\frac{1}{n}\sum_{j=1}^n \m_{t}^j - \nabla f(\x_{t})}^2\right] + \frac{\beta_1^2\sigma^2}{n} + \frac{8L^2\eta^2 d}{\beta_1}.
     \end{split}
\end{align}
Very similarly, we obtain 
\begin{align*}
     \E\left[\left\|\frac{1}{n}\sum_{j=1}^n \m_{t+1}^j - \nabla f(\x_{t+1})\right\|^2\right]\leq (1-\beta_2)\E\left[\Norm{\frac{1}{n}\sum_{j=1}^n \m_{t}^j - \nabla f(\x_{t})}^2\right] + \frac{\beta_2^2\sigma^2}{n} + \frac{8L^2\eta^2 d}{\beta_2}.
\end{align*}
By summing up and rearranging, we observe
\begin{equation}\label{vr}
    \begin{split}
         \E\left[\frac{1}{T}\sum_{t=1}^T \Norm{\frac{1}{n}\sum_{j=1}^n \m_{t}^j - \nabla f(\x_{t})}^2\right]
     \leq &\frac{\E\left[\Norm{\frac{1}{n}\sum_{j=1}^n \m_{1}^j - \nabla f(\x_{1})}^2\right]}{\beta_2 T} + \frac{\beta_2\sigma^2}{n} + \frac{8L^2 \eta^2 d}{\beta_2^2}\\
     \leq &\frac{\sigma^2}{n\beta_2 T} + \frac{\sigma^2 \beta_2}{n} + \frac{8L^2 \eta^2 d}{\beta_2^2}.
    \end{split}
\end{equation}
As a result, by setting that $\beta_2^2 \leq \beta_1 \leq \sqrt{\beta_2}$, we can know that
\begin{align*}
     &\E\left[\frac{1}{T}\sum_{t=1}^T\left\|\frac{1}{n}\sum_{j=1}^n \v_{t}^j - \nabla f(\x_{t})\right\|^2\right]\\
     \leq &\frac{\sigma^2}{nT} + (1-\beta_1)\E\left[\frac{1}{T}\sum_{t=1}^{T-1}\Norm{\frac{1}{n}\sum_{j=1}^n \m_{t}^j - \nabla f(\x_{t})}^2\right] + \frac{\beta_1^2\sigma^2}{n} + \frac{8L^2\eta^2 d}{\beta_1}\\
     \leq & \frac{\sigma^2}{nT} + \frac{\sigma^2}{n\beta_2 T} + \frac{\sigma^2 \beta_2}{n} + \frac{8L^2 \eta^2 d}{\beta_2^2} + \frac{\beta_1^2\sigma^2}{n} + \frac{8L^2\eta^2 d}{\beta_1}\\
     \leq &\frac{2\sigma^2}{\beta_2 n T} + \frac{16\eta^2 L^2 d}{\beta_2^2} + \frac{2\beta_2 \sigma^2}{n}
\end{align*}
By setting that $\beta_2 = \mathcal{O}\left(n^{1/2} T^{-1/2}\right)$, $\eta = \mathcal{O}\left( n^{1/4}d^{-1/2}T^{-3/4}\right)$, and letting $T\ge n$ we observe:
\begin{equation*}
    \begin{split}
        \E\left[\frac{1}{T}\sum_{t=1}^{T} \|\nabla f(\x_t)\|_1\right] 
& \leq \frac{2 \Delta_f}{\eta T} +4\sqrt{d} \cdot \sqrt{\frac{2\sigma^2}{\beta_2 n T} + \frac{16\eta^2 L^2 d}{\beta_2^2} + \frac{2\beta_2 \sigma^2}{n}}+ 4\eta L d\\
& \leq \mathcal{O}\left(\frac{d^{1/2}}{(nT)^{1/4}} \right)
    \end{split}
\end{equation*}

\section{Proof of Theorem~\ref{thm4}}
The analysis of Theorem~\ref{thm1++} is also valid for Theorem~\ref{thm4} until the equation~(\ref{smooth}). The first difference is that we can bound the term $\E\left[\frac{1}{T}\sum_{t=1}^T \Norm{\frac{1}{n}\sum_{j=1}^n \m_{t}^j - \nabla f(\x_{t})}^2\right]$ as follows.

For each worker $j$, we have the following according to the definition of $\m_t^j$:
\begin{align*}
    &\m_{t+1}^j - \nabla f_j(\x_{t+1})
     = (1-\beta_2)\left(\m_t^j - \nabla f_j(\x_t)\right) + \beta_2 \left(\nabla f_j(\x_{t+1};\xi_{t+1}^{j}) - \nabla f_j(\x_{t+1})\right)\\
    & \qquad \qquad+ (1-\beta_2)  \left(\nabla f_j(\x_{t+1};\xi_{t+1}^{j}) - \nabla f_j(\x_{t};\xi_{t+1}^{j}) + \nabla f_j(\x_{t}) - \nabla f_j(\x_{t+1}) \right).
\end{align*}
Summing over $\{n\}$ and noting that $\nabla f(\x) = \frac{1}{n}\sum_{j=1}^n \nabla f_j(\x)$, we can obtain:
\begin{align*}
    &\frac{1}{n}\sum_{j=1}^n \m_{t+1}^j - \nabla f(\x_{t+1}) = \frac{1}{n}\sum_{j=1}^n \left(\m_{t+1}^j - \nabla f_j(\x_{t+1})\right)\\
     = &(1-\beta_2)\frac{1}{n}\sum_{j=1}^n \left(\m_t^j - \nabla f_j(\x_t)\right) + \beta_2 \frac{1}{n}\sum_{j=1}^n \left(\nabla f_j(\x_{t+1};\xi_{t+1}^{j}) - \nabla f_j(\x_{t+1})\right)\\
    & \quad + (1-\beta_2) \frac{1}{n}\sum_{j=1}^n \left(\nabla f_j(\x_{t+1};\xi_{t+1}^{j}) - \nabla f_j(\x_{t};\xi_{t+1}^{j}) + \nabla f_j(\x_{t}) - \nabla f_j(\x_{t+1}) \right).
\end{align*}
Then we have
\begin{align*}
     & \E\left[\left\|\frac{1}{n}\sum_{j=1}^n \m_{t+1}^j - \nabla f(\x_{t+1})\right\|^2\right]\\
     \leq &(1-\beta_2)^2\E\left[\Norm{ \frac{1}{n}\sum_{j=1}^n\left(\m_{t}^j - \nabla f_j(\x_{t})\right)}^2\right]+ 2\beta_2^2 \frac{1}{n^2}\sum_{j=1}^n\E\left[\Norm{ \nabla f_j(\x_{t+1};\xi_{t+1}^{j}) - \nabla f_j(\x_{t+1})}^2\right]\\
     & +2(1-\beta_2)^2 \frac{1}{n^2}\sum_{j=1}^n\E\left[\Norm{ \nabla f_j(\x_{t+1};\xi_{t+1}^{j}) - \nabla f_j(\x_{t};\xi_{t+1}^{j})}^2\right]\\
     \leq &(1-\beta_2)\E\left[\Norm{\frac{1}{n}\sum_{j=1}^n \left(\m_{t}^j - \nabla f_j(\x_{t})\right)}^2\right] + \frac{2\beta_2^2\sigma^2}{n} + \frac{2L^2}{n}\Norm{\x_{t+1}-\x_t}^2\\
     \leq &(1-\beta_2)\E\left[\Norm{\frac{1}{n}\sum_{j=1}^n \m_{t}^j - \nabla f(\x_{t})}^2\right] + \frac{2\beta_2^2\sigma^2}{n} + \frac{8L^2\eta^2 d}{n}.
\end{align*}
By summing up and rearranging, we have
\begin{equation}\label{vr2}
    \begin{split}
         \E\left[\frac{1}{T}\sum_{t=1}^T \Norm{\frac{1}{n}\sum_{j=1}^n \m_{t}^j - \nabla f(\x_{t})}^2\right]
     \leq &\frac{\E\left[\Norm{\frac{1}{n}\sum_{j=1}^n \m_{1}^j - \nabla f(\x_{1})}^2\right]}{\beta_2 T} + \frac{2\sigma^2 \beta_2}{n} + \frac{8L^2 \eta^2 d}{n\beta_2}\\
     \leq &\frac{\sigma^2}{n\beta_2 B_0 T} + \frac{2\sigma^2 \beta_2}{n} + \frac{8L^2 \eta^2 d}{n\beta_2},
    \end{split}
\end{equation}
where $B_0$ is the batch size in the iteration number. Very similarly, we have:
\begin{align*}
     & \E\left[\left\|\frac{1}{n}\sum_{j=1}^n \v_{t+1}^j - \nabla f(\x_{t+1})\right\|^2\right]
     \leq (1-\beta_1)\E\left[\Norm{\frac{1}{n}\sum_{j=1}^n \m_{t}^j - \nabla f(\x_{t})}^2\right] + \frac{2\beta_1^2\sigma^2}{n} + \frac{8L^2\eta^2 d}{n}.
\end{align*}
As a result, by setting that $\beta_1 \leq \sqrt{\beta_2}$, we can know that
\begin{align*}
     &\E\left[\frac{1}{T}\sum_{t=1}^T\left\|\frac{1}{n}\sum_{j=1}^n \v_{t}^j - \nabla f(\x_{t})\right\|^2\right]\\
     \leq &\frac{\sigma^2}{nT} + (1-\beta_1)\E\left[\frac{1}{T}\sum_{t=1}^{T-1}\Norm{\frac{1}{n}\sum_{j=1}^n \m_{t}^j - \nabla f(\x_{t})}^2\right] + \frac{2\beta_1^2\sigma^2}{n} + \frac{8L^2\eta^2 d}{n}\\
     \leq & \frac{\sigma^2}{nT} + \frac{\sigma^2}{n\beta_2 B_0 T} + \frac{2\sigma^2 \beta_2}{n} + \frac{8L^2 \eta^2 d}{n \beta_2} + \frac{2\beta_1^2\sigma^2}{n} + \frac{8L^2\eta^2 d}{n}\\
     \leq &\frac{2\sigma^2}{\beta_2 n B_0 T} + \frac{16\eta^2 L^2 d}{n \beta_2} + \frac{4\beta_2 \sigma^2}{n}
\end{align*}
Set $\beta_2 = \mathcal{O}\left(n^{1/3} T^{-2/3}\right)$, $\eta = \mathcal{O}\left( n^{1/3}d^{-1/2}T^{-2/3}\right)$, $B_0 = \mathcal{O}(n^{-2/3}T^{1/3})$ and $T \ge n^2$ we have:
\begin{equation*}
    \begin{split}
        \E\left[\frac{1}{T}\sum_{t=1}^{T} \|\nabla f(\x_t)\|_1\right] 
& \leq \frac{2 \Delta_f}{\eta T} +4\sqrt{d} \cdot \sqrt{\frac{2\sigma^2}{\beta_2 n B_0 T} + \frac{16\eta^2 L^2 d}{n\beta_2} + \frac{4\beta_2 \sigma^2}{n}}+ 4\eta L d \\
&\leq \mathcal{O}\left(\frac{d^{1/2}}{(nT)^{1/3}} \right)
    \end{split}
\end{equation*}

\section{Proof of Theorem~\ref{thm5}}
Since the overall objective function $f(\x)$ is $L$-smooth, we have the following:
\begin{equation}\label{majorvote}
    \begin{split}
        f(\x_{t+1})\leq& f(\x_t) + \left\langle \nabla f(\x_t), \x_{t+1} - \x_t \right\rangle + \frac{L}{2} \| \x_{t+1} - \x_t \|^2 \\
\leq& f(\x_t) -\eta \left\langle \nabla f(\x_t), \operatorname{Sign}\left(\frac{1}{n}\sum_{j=1}^n \operatorname{S}_G(\v_t^j) \right)\right\rangle -\eta \lambda \left\langle \nabla f(\x_t),  \x_t \right\rangle + 2\eta^2 Ld \\
=& f(\x_t)+  \eta \left\langle \nabla f(\x_t), \sign(\nabla f(\x_t))-\operatorname{Sign}\left(\frac{1}{n}\sum_{j=1}^n \operatorname{S}_G(\v_t^j) \right)\right\rangle\\
&\quad - \eta\left\langle \nabla f(\x_t), \sign(\nabla f(\x_t)) \right\rangle + \eta \lambda \Norm{\nabla f(\x_t)}_1\Norm{\x_t}_\infty + 2\eta^2 Ld \\
=& f(\x_t)+  \eta \left\langle \nabla f(\x_t), \sign(\nabla f(\x_t))-\operatorname{Sign}\left(\frac{1}{n}\sum_{j=1}^n \operatorname{S}_G(\v_t^j) \right)\right\rangle-\frac{\eta}{2}  \Norm{ \nabla f(\x_t)}_1 + 2\eta^2 Ld \\
\leq& f(\x_t)+ 2\eta G\sqrt{d}   \Norm{\frac{\nabla f(\x_t)}{G} - \frac{1}{n}\sum_{j=1}^n \operatorname{S}_G(\v_t^j)} - \frac{\eta}{2}  \Norm{ \nabla f(\x_t)}_1+2\eta^2 Ld,
    \end{split}
\end{equation}
where the last inequality is because of
\begin{equation}\label{equality2}
\begin{aligned}
&\left\langle \nabla f(\x_t), \sign(\nabla f(\x_t)) - \operatorname{Sign}\left(\frac{1}{n}\sum_{j=1}^n \operatorname{S}_G(\v_t^j) \right) \right\rangle \\
= &\sum_{i=1}^{d} \left\langle [\nabla f(\x_t)]_i, \sign([\nabla f(\x_t)]_i) - \sign\left(\left[\frac{1}{n}\sum_{j=1}^n \operatorname{S}_G(\v_t^j)\right]_i\right) \right\rangle  \\
\leq &\sum_{i=1}^{d} 2G\left|\frac{[\nabla f(\x_t)]_i }{G}\right| \cdot \mathbb{I} \left( \sign(\left[\nabla f(\x_t)\right]_i) \neq \sign\left(\left[\frac{1}{n}\sum_{j=1}^n \operatorname{S}_G(\v_t^j)\right]_i\right) \right) \\
\leq & \sum_{i=1}^{d} 2G\left|\frac{[\nabla f(\x_t)]_i}{G} - \left[\frac{1}{n}\sum_{j=1}^n \operatorname{S}_G(\v_t^j)\right]_i\right| \cdot \mathbb{I} \left( \sign(\left[\nabla f(\x_t)\right]_i) \neq \sign\left(\left[\frac{1}{n}\sum_{j=1}^n \operatorname{S}_G(\v_t^j)\right]_i\right) \right) \\
\leq &\sum_{i=1}^{d} 2G\left|\frac{[\nabla f(\x_t)]_i}{G} - \left[\frac{1}{n}\sum_{j=1}^n \operatorname{S}_G(\v_t^j)\right]_i\right| \\
= & 2G\Norm{\frac{\nabla f(\x_t)}{G} -\frac{1}{n}\sum_{j=1}^n \operatorname{S}_G(\v_t^j)}_1 \leq 2G\sqrt{d} \Norm{\frac{\nabla f(\x_t)}{G} - \frac{1}{n}\sum_{j=1}^n \operatorname{S}_G(\v_t^j)}.
\end{aligned}    
\end{equation}
Rearranging and taking the expectation, we have:
\begin{equation}\label{majorvote2}
    \begin{split}
&\E\left[ \Norm{\frac{\nabla f(\x_t)}{G} - \frac{1}{n}\sum_{j=1}^n \operatorname{S_\textit{G}}(\v_t^j)} \right] \\
\leq &   \E\left[\Norm{ \frac{\nabla f(\x_t)}{G} - \frac{1}{nG}\sum_{j=1}^n \v_t^j}\right]+  \E\left[  \Norm{ \frac{1}{n}\sum_{j=1}^n \left(\operatorname{S_\textit{G}}(\v_t^j)-\frac{\v_t^j}{G}\right)} \right]\\
\leq &  \E\left[\frac{1}{G}\Norm{ \nabla f(\x_t) - \frac{1}{n}\sum_{j=1}^n \v_t^j}\right]+ \sqrt{\E\left[  \Norm{ \frac{1}{n}\sum_{j=1}^n \left(\operatorname{S_\textit{G}}(\v_t^j)-\frac{\v_t^j}{G}\right)}^2 \right]}\\
\leq & \E\left[\frac{1}{G}\Norm{ \nabla f(\x_t) - \frac{1}{n}\sum_{j=1}^n \v_t^j}\right]+ \sqrt{\frac{1}{n^2}\sum_{j=1}^n\E\left[  \Norm{  \left(\operatorname{S_\textit{G}}(\v_t^j)-\frac{\v_t^j}{G}\right)}^2 \right]}\\
\leq &  \E\left[\frac{1}{G}\Norm{ \nabla f(\x_t) - \frac{1}{n}\sum_{j=1}^n \v_t^j}\right]+  \sqrt{\frac{1}{n^2}\sum_{j=1}^n\E\left[  \Norm{ \operatorname{S_\textit{G}}(\v_t^j)}^2 \right]}\\
\leq &  \E\left[\frac{1}{G}\Norm{ \nabla f(\x_t) - \frac{1}{n}\sum_{j=1}^n \v_t^j}\right]+ \frac{\sqrt{d}}{\sqrt{n}}, 
    \end{split}
\end{equation}
where the second inequality is due to the fact that $\left(\E\left[X\right]\right)^2 \leq \E\left[X^2\right]$, and the forth inequality is because of $\E\left[S_G\left(\v_t^j\right) \right] = \frac{\v_t^j}{G}$, as well as the $S_G$ operation in each node is independent.

Rearranging the terms and summing up, we have:
\begin{align*}
     \frac{1}{T} \sum_{i=1}^{T} \E \left[\left\| \nabla f(\x_t) \right\|_1 \right]&\leq \frac{2\Delta_f}{\eta T} +4\sqrt{d}\E\left[\frac{1}{T} \sum_{i=1}^{T} \left\| \nabla f(\x_t) - \frac{1}{n}\sum_{j=1}^{n}\v_t^j  \right\|\right]+ \frac{4d G}{\sqrt{n}} + 4\eta L d \\
     &\leq \frac{2\Delta_f}{\eta T} +4\sqrt{d}\sqrt{\E\left[\frac{1}{T} \sum_{i=1}^{T} \left\| \nabla f(\x_t) - \frac{1}{n}\sum_{j=1}^{n}\v_t^j  \right\|^2 \right]}+ \frac{4d G}{\sqrt{n}} + 4\eta L d,
\end{align*}
where the last inequality is due to Jensen's inequality.

Note that in Theorem~\ref{thm3}, we have already shown that 
\begin{align*}
     \E\left[\frac{1}{T}\sum_{t=1}^T\left\|\frac{1}{n}\sum_{j=1}^n \v_{t}^j - \nabla f(\x_{t})\right\|^2\right] \leq &\frac{2\sigma^2}{\beta_2 n T} + \frac{16\eta^2 L^2 d}{\beta_2^2} + \frac{2\beta_2 \sigma^2}{n}
\end{align*}
Finally, we can ensure that
\begin{align*}
     \frac{1}{T} \sum_{i=1}^{T} \E \left[\| \nabla f(\x_t) \|_1 \right]&\leq \frac{2\Delta_f}{\eta T} + \frac{4d G}{\sqrt{n}} +4\eta L d+4\sqrt{d}\sqrt{\E\left[\frac{1}{T} \sum_{i=1}^{T} \left\| \nabla f(\x_t) - \frac{1}{n}\sum_{j=1}^{n}\v_t^j  \right\|^2 \right]}\\
     & \leq \frac{2\Delta_f}{\eta T}+  \frac{4d G}{\sqrt{n}} + 4\eta L d  + 4\sqrt{d}\sqrt{\frac{2\sigma^2}{n\beta_2 T} + \frac{2\sigma^2 \beta_2}{n} + \frac{16L^2 \eta^2 d}{\beta_2^2}}.
\end{align*}
By setting $\beta_2=\frac{1}{2}$ and $\eta = \mathcal{O}\left({T^{-1/2}d^{-1/2}}\right)$, we have
\begin{align*}
     \frac{1}{T} \sum_{i=1}^{T} \| \nabla f(\x_t) \|_1 =\mathcal{O}\left( \frac{d^{1/2}}{T^{1/2}} + \frac{d}{n^{1/2}} \right).
\end{align*}
By setting $\beta_2=\frac{1}{2}$ and $\eta = \mathcal{O}\left({n^{-1/2}}\right)$, we have
\begin{align*}
     \frac{1}{T} \sum_{i=1}^{T} \| \nabla f(\x_t) \|_1 =\mathcal{O}\left( \frac{n^{1/2}}{T} + \frac{d}{n^{1/2}} \right).
\end{align*}

\section{Proof of Theorem~\ref{thm7}}
Since function $f(\x)$ is $L$-smooth, we have the following by setting $\lambda \leq \frac{\sqrt{L}}{T\sqrt{\eta G}}$:
\begin{equation*}
    \begin{split}
        f(\x_{t+1})\leq& f(\x_t) + \left\langle \nabla f(\x_t), \x_{t+1} - \x_t \right\rangle + \frac{L}{2} \| \x_{t+1} - \x_t \|^2 \\
\leq& f(\x_t) -\eta \left\langle \nabla f(\x_t), \operatorname{S_1}\left(\frac{1}{n}\sum_{j=1}^n \operatorname{S_\textit{G}}({\v}_t^j) \right)\right\rangle -\eta \lambda \left\langle \nabla f(\x_t),  \x_t \right\rangle + 2\eta^2 Ld\\
=& f(\x_t)+  \eta \left\langle \nabla f(\x_t),\frac{\nabla f(\x_t)}{G} -\operatorname{S_1}\left(\frac{1}{n}\sum_{j=1}^n \operatorname{S_\textit{G}}({\v}_t^j) \right)\right\rangle  \\
&\quad - \eta \left\langle \nabla f(\x_t),\frac{\nabla f(\x_t)}{G} \right\rangle + \frac{\eta G}{2} \lambda^2 \Norm{\x_t}^2+ \frac{\eta }{2G}\Norm{\nabla f(\x_t)}^2 + 2\eta^2 Ld \\
=& f(\x_t)+  \eta \left\langle \nabla f(\x_t),\frac{\nabla f(\x_t)}{G} -\operatorname{S_1}\left(\frac{1}{n}\sum_{j=1}^n \operatorname{S_\textit{G}}({\v}_t^j) \right)\right\rangle  - \frac{\eta}{2G}\Norm{\nabla f(\x_t)}^2 + 3\eta^2 Ld.
    \end{split}
\end{equation*}
Taking expectations leads to:
\begin{equation}\label{majorvote3}
    \begin{split}
       &\E\left[ f(\x_{t+1})-f(\x_t) \right]\\
       \leq&  \eta \E\left[\left\langle \nabla f(\x_t), \frac{1}{G}\nabla f(\x_t)-\operatorname{S_1}\left(\frac{1}{n}\sum_{j=1}^n \operatorname{S_\textit{G}}({\v}_t^j) \right)\right\rangle \right] - \frac{\eta}{2G}\E\left[\Norm{\nabla f(\x_t)}^2\right] + 3\eta^2 Ld\\
       =&  \eta \E\left[\left\langle \nabla f(\x_t), \frac{1}{G}\nabla f(\x_t)-\frac{1}{n}\sum_{j=1}^n \operatorname{S_\textit{G}}({\v}_t^j) \right\rangle \right] - \frac{\eta}{2G}\E\left[\Norm{\nabla f(\x_t)}^2\right] + 3\eta^2 Ld\\
       =&  \eta \E\left[\left\langle \nabla f(\x_t), \frac{1}{G}\nabla f(\x_t)-\frac{1}{nG}\sum_{j=1}^n {\v}_t^j \right\rangle \right] - \frac{\eta}{2G}\E\left[\Norm{\nabla f(\x_t)}^2\right] + 3\eta^2 Ld\\
       \leq&  \eta \E\left[\frac{1}{4G}\Norm{\nabla f(\x_t)}^2+\frac{1}{G}\Norm{\nabla f(\x_t) -\frac{1}{n}\sum_{j=1}^n {\v}_t^j}^2  \right] - \frac{\eta}{2G}\E\left[\Norm{\nabla f(\x_t)}^2\right] + 3\eta^2 Ld\\
       =&  \frac{\eta}{G} \E\left[\Norm{\nabla f(\x_t) -\frac{1}{n}\sum_{j=1}^n {\v}_t^j}^2  \right] - \frac{\eta}{4G}\E\left[\Norm{\nabla f(\x_t)}^2\right] + 3\eta^2 Ld.
    \end{split}
\end{equation}
Rearranging the terms and summing up:
\begin{align*}
     \frac{1}{T} \sum_{i=1}^{T} \E\left[\| \nabla f(\x_t) \|^2 \right]&\leq \frac{4\Delta_f G}{\eta T} +4\E\left[\frac{1}{T} \sum_{i=1}^{T} \left\| \nabla f(\x_t) - \frac{1}{n}\sum_{j=1}^{n}{\v}_t^j  \right\|^2\right] + 12\eta L d G
\end{align*}
Note that in Theorem~\ref{thm3}, we have already shown that 
\begin{align*}
     \E\left[\frac{1}{T}\sum_{t=1}^T\left\|\frac{1}{n}\sum_{j=1}^n \v_{t}^j - \nabla f(\x_{t})\right\|^2\right] \leq &\frac{2\sigma^2}{\beta_2 n T} + \frac{16\eta^2 L^2 d}{\beta_2^2} + \frac{2\beta_2 \sigma^2}{n}
\end{align*}
Finally, we can obtain the final bound:
\begin{align*}
     \E\left[ \frac{1}{T} \sum_{i=1}^{T} \| \nabla f(\x_t) \| \right]& \leq \sqrt{\E\left[ \frac{1}{T} \sum_{i=1}^{T} \| \nabla f(\x_t) \|^2 \right]} \\
     & \leq \sqrt{ \frac{4\Delta_f G}{\eta T}+ 12\eta L d G  +\frac{8\sigma^2}{\beta_2 n T} + \frac{8\sigma^2 \beta_2}{n} + \frac{64L^2 \eta^2 d}{\beta_2^2}}.
\end{align*}
That is to say, by setting $\beta_2 = n^{1/3} \eta^{2/3} d^{1/3}$, $\eta = \mathcal{O}\left(\min \left\{\frac{1}{T^{1/2}d^{1/2} }, \frac{n^{2/5}}{T^{3/5}d^{1/5}}\right\}\right)$, we can obtain the convergence rate of $\mathcal{O}\left(\max \left\{\frac{d^{1/4}}{T^{1/4}},\frac{d^{1/10}}{n^{1/5}T^{1/5}} \right\} \right)$.

\section{Proof of Theorem~\ref{thm8}}
According to the previous analysis of Theorem~\ref{thm7}, we have that
\begin{align*}
     \frac{1}{T} \sum_{i=1}^{T} \E\left\| \nabla f(\x_t) \|^2 \right]&\leq \frac{4\Delta_f G}{\eta T} +4\E\left[\frac{1}{T} \sum_{i=1}^{T} \left\| \nabla f(\x_t) - \frac{1}{n}\sum_{j=1}^{n}{\v}_t^j  \right\|^2\right] + 12\eta L d G
\end{align*}
Using the same analysis as in equation~(\ref{SSR}), we can ensure that
\begin{align*}
\begin{split}
     \E\left[\left\|\frac{1}{n}\sum_{j=1}^n \v_{t+1}^j - \nabla f(\x_{t+1})\right\|^2\right]
     \leq &(1-\beta_1)\E\left[\Norm{\frac{1}{n}\sum_{j=1}^n \m_{t}^j - \nabla f(\x_{t})}^2\right] + \frac{\beta_1^2\sigma^2}{n} + \frac{8L^2\eta^2 d}{\beta_1}.
     \end{split}
\end{align*}
Also, in equation~(\ref{vr2}), we have already shown that
\begin{equation*}
    \begin{split}
         \E\left[\frac{1}{T}\sum_{t=1}^T \Norm{\frac{1}{n}\sum_{j=1}^n \m_{t}^j - \nabla f(\x_{t})}^2\right]
     \leq \frac{\sigma^2}{n\beta_2 B_0 T} + \frac{2\sigma^2 \beta_2}{n} + \frac{8L^2 \eta^2 d}{n\beta_2},
    \end{split}
\end{equation*}
As a result, by setting that $\beta_1 \leq \sqrt{\beta_2}$, we can know that
\begin{align*}
     &\E\left[\frac{1}{T}\sum_{t=1}^T\left\|\frac{1}{n}\sum_{j=1}^n \v_{t}^j - \nabla f(\x_{t})\right\|^2\right]\\
     \leq &\frac{\sigma^2}{nT} + (1-\beta_1)\E\left[\frac{1}{T}\sum_{t=1}^{T-1}\Norm{\frac{1}{n}\sum_{j=1}^n \m_{t}^j - \nabla f(\x_{t})}^2\right] + \frac{\beta_1^2\sigma^2}{n} + \frac{8L^2\eta^2 d}{\beta_1}\\
     \leq & \frac{\sigma^2}{nT} + \frac{\sigma^2}{n\beta_2 B_0 T} + \frac{2\sigma^2 \beta_2}{n} + \frac{8L^2 \eta^2 d}{n \beta_2} + \frac{\beta_1^2\sigma^2}{n} + \frac{8L^2\eta^2 d}{\beta_1}\\
     \leq &\frac{2\sigma^2}{\beta_2 n B_0 T} + \frac{16\eta^2 L^2 d}{\beta_2} + \frac{4\beta_2 \sigma^2}{n}
\end{align*}
Finally, we can obtain the final bound:
\begin{align*}
     \E\left[ \frac{1}{T} \sum_{i=1}^{T} \| \nabla f(\x_t) \| \right]& \leq \sqrt{\E\left[ \frac{1}{T} \sum_{i=1}^{T} \| \nabla f(\x_t) \|^2 \right]} \\
     & \leq \sqrt{ \frac{4\Delta_f G}{\eta T}+ 12\eta L d G + \frac{2\sigma^2}{\beta_2 n B_0 T} + \frac{16\eta^2 L^2 d}{\beta_2} + \frac{4\beta_2 \sigma^2}{n}}.
\end{align*}
That is to say, by setting $\beta_2 = \mathcal{O}\left(\frac{1}{T^{1/2}}\right)$, $\eta = \mathcal{O}\left(\frac{1}{T^{1/2}d^{1/2} }\right)$, we can obtain the convergence rate of $\mathcal{O}\left(\frac{d^{1/4}}{T^{1/4}} \right)$.
\end{document}